%% file: main.tex
\documentclass[letterpaper, 10 pt, conference]{ieeeconf}  

\usepackage{array}
\usepackage[caption=false,font=footnotesize,labelfont=sf,textfont=sf]{subfig}
\usepackage{textcomp}
\usepackage{stfloats}
\usepackage{url}
\usepackage{verbatim}
\usepackage{graphicx}
\usepackage{amssymb}
\usepackage{pifont}
\usepackage{siunitx}
\usepackage{algorithm,algcompatible}
\usepackage{todonotes}
\usepackage{hyperref}
\usepackage[capitalize]{cleveref}
\usepackage{comment}
\usepackage{multirow}
\usepackage{tikz}
\usepackage{pgfplots}
\usepackage{caption}
\usepackage{cite}
\usepackage{tabularx}
\usepackage{booktabs}
\usepackage{float}

\usepackage{xcolor}
\definecolor{Bluelight}{HTML}{0065BD} 
\definecolor{Black}{HTML}{000000}
\definecolor{Blue}{HTML}{005293}
\definecolor{Bluestrong}{HTML}{003359}
\definecolor{Red}{HTML}{8C000F}
\definecolor{Grey}{HTML}{808080}
\definecolor{Greylight}{HTML}{CCCCCC}
\definecolor{Orange}{HTML}{E37222}
\definecolor{Green}{HTML}{A2AD00}
\definecolor{GreenCR}{HTML}{008000}
\definecolor{OrangeCR}{HTML}{f1b514}
\definecolor{visibleArea}{HTML}{81AF82}
\definecolor{occludedAreaObst}{HTML}{DC8282}
\definecolor{occludedAreaLanelet}{HTML}{DBBC82}

\algnewcommand\FUNC{\item[\textbf{Function Signature:}]}
\algnewcommand\INPUT{\item[\textbf{Input:}]}
\algnewcommand\OUTPUT{\item[\textbf{Output:}]}
\algnewcommand\RETURN{\STATE{\textbf{return\ }}}

\IEEEoverridecommandlockouts                              

\title{\LARGE \bf
MIND-Stack: Modular, Interpretable, End-to-End Differentiability for Autonomous Navigation}

\author{Felix Jahncke, Johannes Betz 
\thanks{The code is provided open-source and available via the following link: \url{https://github.com/TUM-AVS/MIND-Stack}}
\thanks{F. Jahncke and J. Betz are with the Professorship of Autonomous Vehicle Systems, TUM School of Engineering and Design, Technical University Munich, 85748 Garching, Germany; Munich Institute of Robotics and Machine Intelligence (MIRMI), \{{felix.jahncke}\}@tum.de}
}

\newcommand\copyrighttext{%
  \footnotesize \textcopyright 2025 IEEE.  Personal use of this material is permitted.  Permission from IEEE must be obtained for all other uses, in any current or future media, including reprinting/republishing this material for advertising or promotional purposes, creating new collective works, for resale or redistribution to servers or lists, or reuse of any copyrighted component of this work in other works. 
  } 
\newcommand\copyrightnotice{%
\begin{tikzpicture}[remember picture,overlay]
\node[anchor=south,yshift=10pt] at (current page.south) {\fbox{\parbox{\dimexpr\textwidth-\fboxsep-\fboxrule\relax}{\copyrighttext}}};
\end{tikzpicture}%
}

\begin{document}

\maketitle

\copyrightnotice

\begin{abstract}
Developing robust, efficient navigation algorithms is challenging. Rule-based methods offer interpretability and modularity but struggle with learning from large datasets, while end-to-end neural networks excel in learning but lack transparency and modularity.
In this paper, we present MIND-Stack,  a modular software stack consisting of a localization network and a Stanley Controller with intermediate human interpretable state representations and end-to-end differentiability.
Our approach enables the upstream localization module to reduce the downstream control error, extending its role beyond state estimation.
Unlike existing research on differentiable algorithms that either lack modules of the autonomous stack to span from sensor input to actuator output or real-world implementation, MIND-Stack offers both capabilities.
We conduct experiments that demonstrate the ability of the localization module to reduce the downstream control loss through its end-to-end differentiability while offering better performance than state-of-the-art algorithms. 
We showcase sim-to-real capabilities by deploying the algorithm on a real-world embedded autonomous platform with limited computation power and demonstrate simultaneous training of both the localization and controller towards one goal.
While MIND-Stack shows good results, we discuss the incorporation of additional modules from the autonomous navigation pipeline in the future, promising even greater stability and performance in the next iterations of the framework.
\end{abstract}

\vspace{0.1cm}

\section{INTRODUCTION}
Developing efficient and well-performing autonomy algorithms for complex navigation applications has been a significant challenge over the past years. 
    Despite continuous advancements, the performance still falls considerably short of the expected developments, integration, and real-world deployment. 
    Currently, two approaches have formed to tackle these challenges: \textit{rule-based modular algorithms} and \textit{neural-network-based end-to-end approaches}.

    Rule-based modular approaches typically subdivide the overall system into task-defined submodules such as perception, planning, and control \cite{machines5010006,basye1992decision}.
    This structure offers advantages such as verifiability and interpretability for humans but often leads to encapsulated modules that are only optimized toward their immediate neighboring modules. 
    In contrast, neural network-based end-to-end algorithms \cite{grigorescuSurveyDeepLearning2020a, chibRecentAdvancementsEndtoEnd2024, kendall2019learning}  offer the capability to learn from large datasets directly regarding the final driving task. 
    A significant challenge is their representation of the entire software stack as a black box and the algorithm's subsequent difficulty in verification and interpretability for humans.

    Recently, new ideas emerged to combine the strengths of both approaches by maintaining the modularity and interpretability of rule-based systems, utilizing hand-crafted modules, and enabling end-to-end differentiability \cite{karkusDiffStackDifferentiableModular2022}.
    Inspired by the idea of a differentiable software stack, we introduce a \textbf{M}odular, \textbf{IN}terpretable, and \textbf{D}ifferentiable Software Stack (\textbf{MIND}-Stack) in this paper. 
    The stack enables end-to-end trainability and optimization of high-level modules based on downstream losses while remaining modular and offering user interpretability from intermediate state representations. 

    \begin{figure}[!t]
        \centering
        \includegraphics[width=1.0\linewidth]{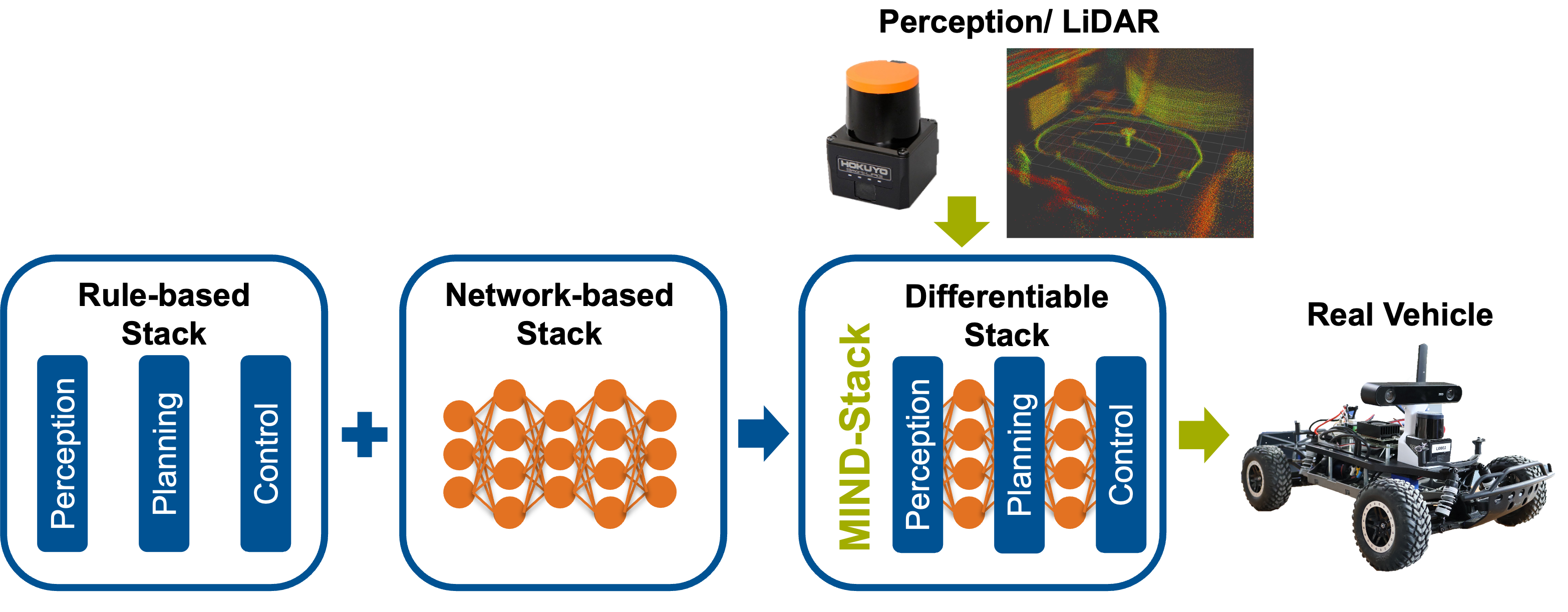}
        \caption{MIND-Stack combines the advantages of rule-based and end-to-end approaches, optimizing the full-autonomy stack from sensor input to vehicle control.}
        \label{fig:enter-label}
    \end{figure}
    
    We explicitly show the capabilities by combining the modules of localization and control, training and evaluating the network in multiple scenarios, and optimizing the control losses. 
    The software stack performs better than state-of-the-art algorithms in simulation, and we demonstrate simulation to real-world transferability on a real-world embedded autonomous platform, as well as combined localization and control training. 
    Our MIND-Stack framework offers the following three contributions:
    \begin{enumerate}
        \item \textbf{Differentiability, modularity, and interpretability}: MIND-Stack overcomes the limitations of rule-based and end-to-end learning approaches by offering a modular architecture and interpretability from intermediate state representations comprehensible to humans, all while remaining fully end-to-end differentiable.
        \item \textbf{Advancing beyond previous research}: Where previous research does not implement a full autonomous driving stack from sensor input to actuator output or lacks real-world implementation, we design MIND-Stack to provide both features.
        \item \textbf{Adaptability and real-world performance}: We demonstrate the capabilities of MIND-Stack in various scenarios, highlighting the abilities of the algorithm to improve performance across various challenges, including the real world, where it effectively handles real-world uncertainties and operates within computational constraints.
    \end{enumerate}

\section{RELATED WORK}

    MIND-Stack is located at the intersection of neural network-based end-to-end and traditional rule-based algorithms. Current approaches achieve differentiability in their software architecture in various ways and degrees. 
    Hybrid models combine neural networks with traditional algorithms for specific parts of their code. 
    By now, neural network modules exist for all parts of the autonomous stack \cite{grigorescuSurveyDeepLearning2020a}, including perception \cite{redmonYouOnlyLook2016, yinCenterbased3DObject2021, liDeepFusionLidarCameraDeep2022}, prediction \cite{wuMotionNetJointPerception2020,mozaffariDeepLearningBasedVehicle2022,houstonOneThousandOne2020}, planning \cite{baiDeepLearningBased2019,klimkeIntegrationReinforcementLearning2023}, and control \cite{salzmannRealTimeNeuralMPC2023, tianParallelLearningBasedSteering2023, dangEventTriggeredModelPredictive2024,NEURIPS2020_5a7b238b}. 
    Full end-to-end networks, with the most established variants of imitation- \cite{caiCarlLeadLidarbasedEndtoEnd2021, chenLearningAllVehicles2022, xiaoMultimodalEndtoEndAutonomous2019,chittaTransFuserImitationTransformerBased2022} and reinforcement learning \cite{toromanoffEndtoEndModelFreeReinforcement2020,chenInterpretableEndtoendUrban2020,stachowiczFastRLAPSystemLearning2023,evansHighspeedAutonomousRacing2023}, apply full differentiability from sensor input to control outputs \cite{chibRecentAdvancementsEndtoEnd2024}. 
    
    Rule-based modular stacks are inherently interpretable, with each module's in-, output, and inner functionality fully transparent and interpretable to humans. 
    The black-box nature of full neural network-based algorithms is a continuous area of research under the term Explainable Artificial Intelligence (AI) \cite{omeizaExplanationsAutonomousDriving2022, atakishiyevExplainableArtificialIntelligence2024}.
    While Explainable AI techniques, such as visual attention models for self-driving cars \cite{kimInterpretableLearningSelfDriving2017}, introduce methods to improve these transparency issues, they do not achieve the same level of interpretability as rule-based architectures.

    \begin{figure*}[ht!]
    \centering
    \begin{minipage}[t]{0.3\textwidth} 
        \centering
        \includegraphics[width=\textwidth]{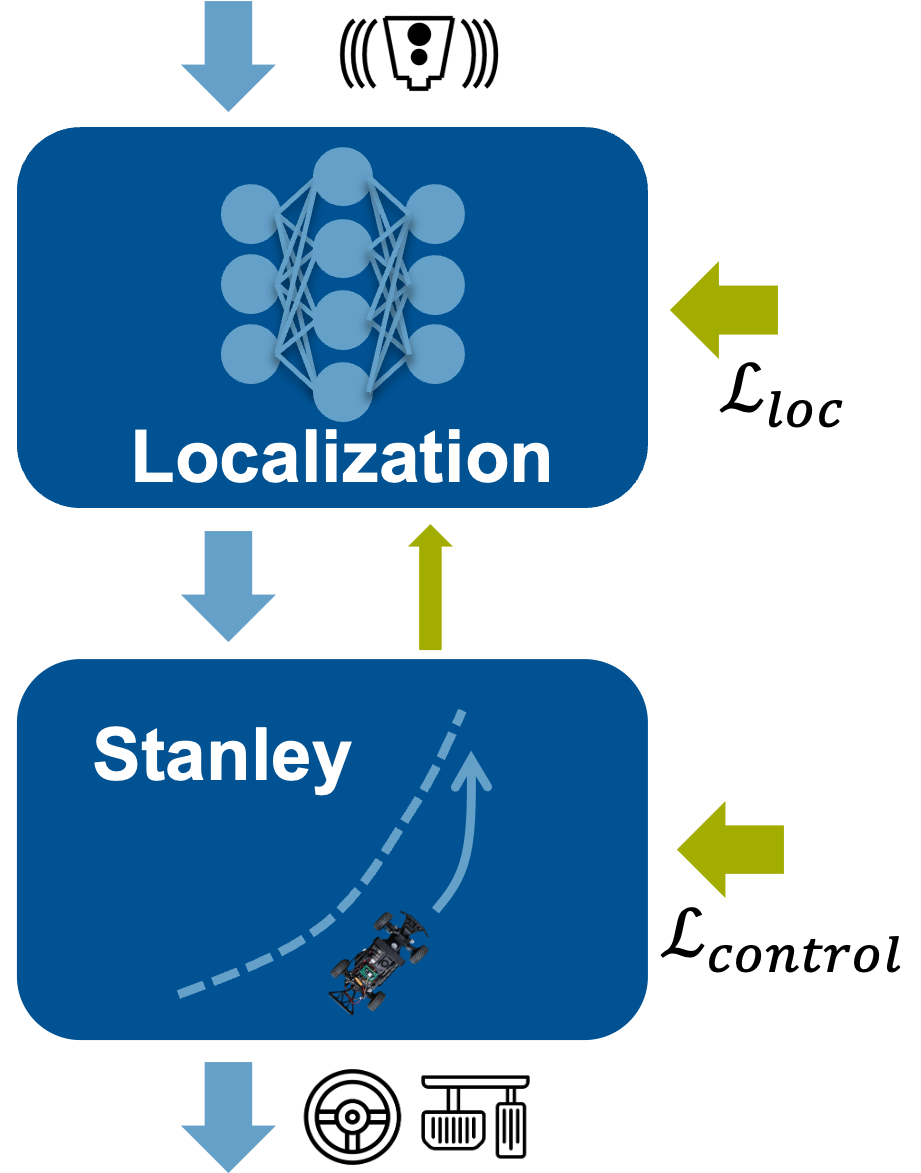}
    \end{minipage}
    \hspace{0.0\textwidth}
    \begin{minipage}[t]{0.3\textwidth}
        \centering
        \input{Plots/Sim_Heading_Loss_Squared/sim_heading_loss_squared}
    \end{minipage}
    \hfill
    \begin{minipage}[t]{0.28\textwidth}
        \centering
        \begin{tikzpicture}
        \node[anchor=south west,inner sep=0] (image) at (0,0) {
            \includegraphics[width=\textwidth]{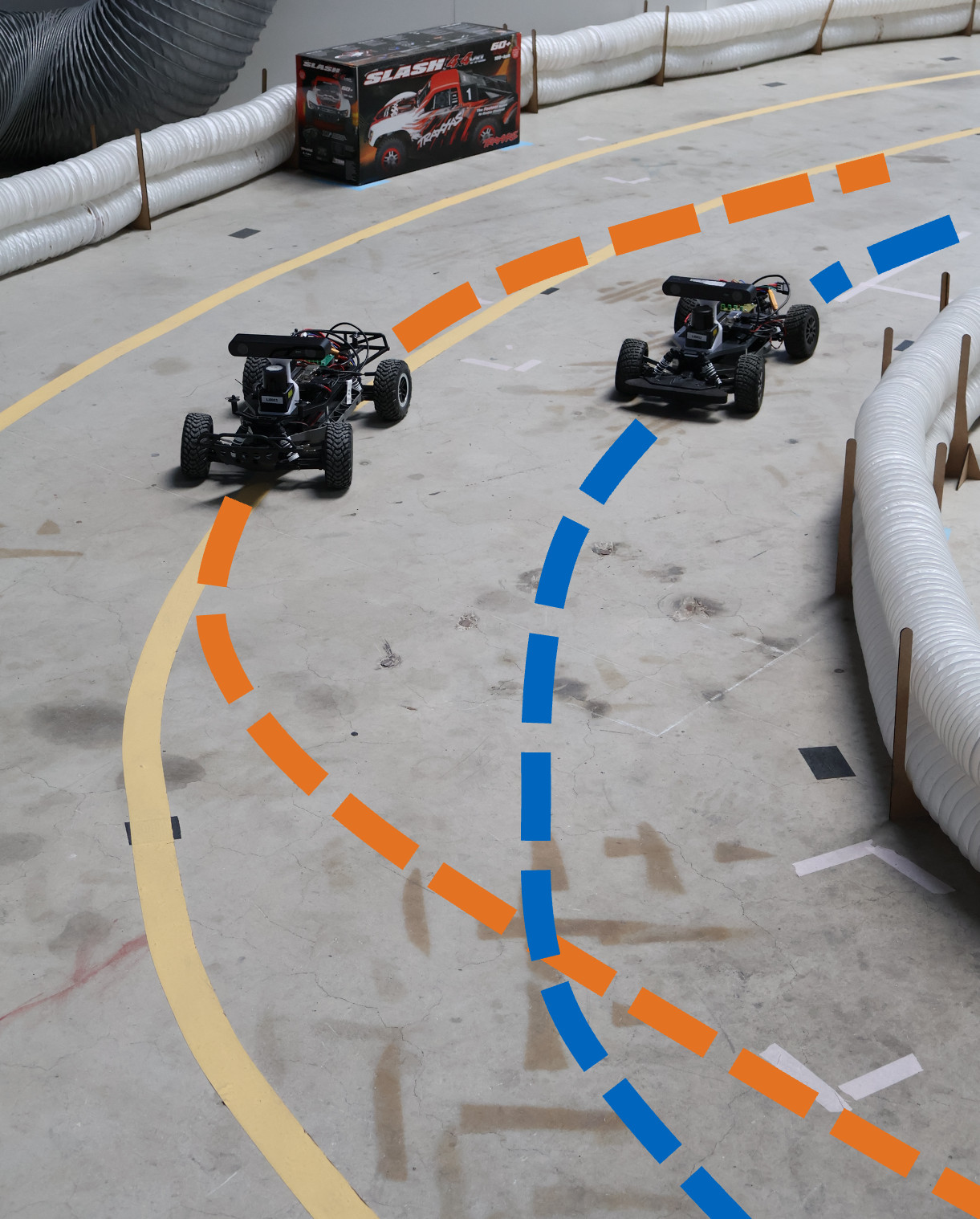}
        };
        \begin{scope}[x={(image.south east)},y={(image.north west)}]
            \node[fill=white,draw=none,anchor=north west, text width=\textwidth] at (-0.01,1.10) {
                \begin{tabular}{@{}rl@{}}
                    \tikz \draw[Bluelight, line width=2.4pt, dashed] (0,0) -- ++(1.65em,0); & \hspace{-1.5em} Traj. before optimization \\
                    \tikz \draw[Orange, line width=2.4pt, dashed] (0,0) -- ++(1.65em,0); & \hspace{-1.5em} Traj. after optimization
                \end{tabular}
            };
        \end{scope}
        \end{tikzpicture}
    \end{minipage}
    \caption{By leveraging a \textbf{modular architecture with end-to-end differentiability}, MIND-Stack enables the upstream localization to improve the downstream control loss (left). 
    MIND-Stack optimizes losses (middle) while being lightweight and efficient, as verified on an autonomous platform (right), where the vehicle learns to optimize its driving policy and trajectory.}
    \label{fig:paper overview}
    \end{figure*}
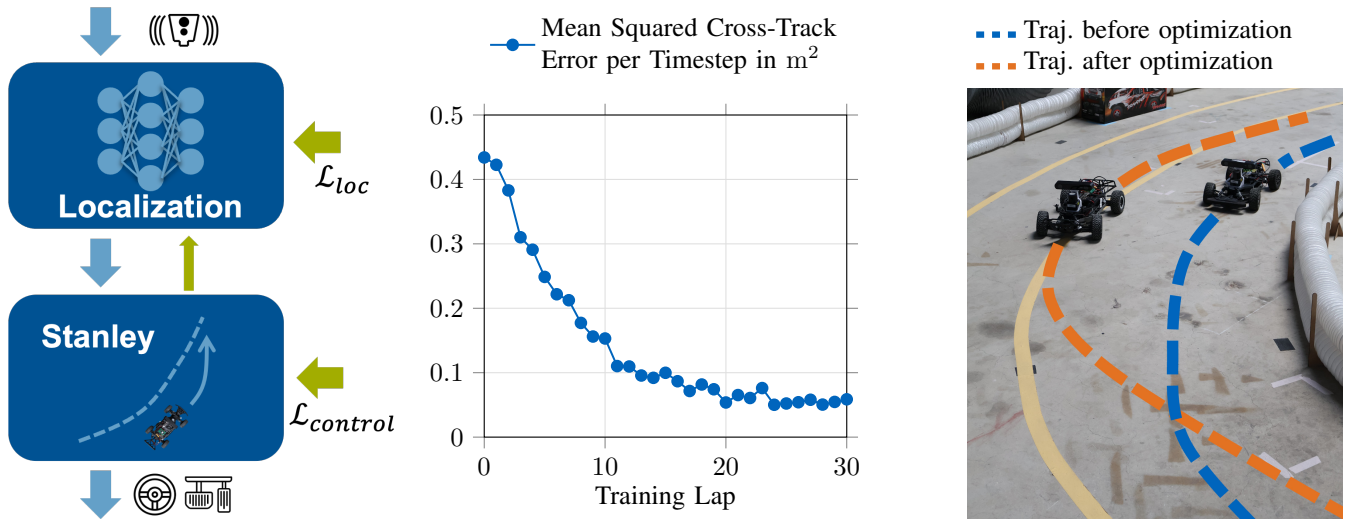

    Modularity involves the systematic algorithm design to include interchangeable components for flexibility and scalability.
    Traditional algorithms with subdivided modules allow for targeted optimization and exchange of modules. 
    Recent research wants to break down AI algorithms into smaller exchangeable systems. 
    The current ideation of such Composable AI \cite{yangDeepModelReassembly2022} has not been evaluated for autonomous driving, and this approach does not allow for incorporating hand-crafted modules.

    While initial ideations of differential software stack implementations exist, DiffStack proposed by Karkus et al. \cite{karkusDiffStackDifferentiableModular2022}, Differentiable Algorithm Networks (DAN) \cite{Lozano-Perez-RSS-19} and PyPose \cite{wangPyPoseLibraryRobot2023, zhan2023pypose} are not without limitation and have fundamental architectural differences that prevent a direct comparison with our work.
    
    DiffStack lacks both a perception and localization module, relying on the ground truth annotations from a dataset, and does not generate vehicle actuator control commands.
    DAN only operates in simplified, grid-based simulation environments and outputs high-level actions rather than actuator control commands.
    Both system additionally rely on ground-truth observations and lack real world verification.
    PyPose similarly lacks perception components and assumes that sufficiently accurate localization is already available on the robot.
    It remains limited to demonstration on legged robots and has not shown maturity for broad applications. 
    In contrast, MIND-Stack offers full end-to-end differentiability from sensor input to actuator output and we demonstrate the advantages of our stack in both simulation and real-world environments.

\section{METHODOLOGY}

    In order to demonstrate the possibilities of our MIND-Stack framework, we combine both localization and lateral control (Stanley Controller \cite{hoffmannAutonomousAutomobileTrajectory2007}) modules into a single, end-to-end, differentiable software stack (Figure \ref{fig:paper overview}).
    This stack enables an autonomous platform to follow a predefined trajectory autonomously, created offline using the trajectory generator proposed by Heilmeier et al. \cite{heilmeierMinimumCurvatureTrajectory2020}.
    We directly use the velocity profile from the generated trajectory and achieve localization with a CNN-based neural network (Section \ref{sec:localization_module}).
    The control module is hand-crafted and optimized to follow the waypoints as closely as possible (Section \ref{sec:control_module}).
    One of the main contributions of MIND-Stack is not in learning the initial state estimator but in optimizing the localization network further using a combination of localization and control losses (Section \ref{sec:end-to-end-training}), specifically to improve downstream control performance.
    The stack enables interpretability through human-interpretable interfaces and, in addition, modularity by separating tasks into distinct modules (Section \ref{sec:mod-and-interpret}).
    
    \subsection{Localization module}
    \label{sec:localization_module}

    We utilize a 1D Convolutional Neural Network (CNN), which takes the current LiDAR scan and the previous vehicle pose as input, and predicts the current global vehicle pose $(\hat{x}, \hat{y}, \hat{\theta})$ as output.

    The architecture of the CNN is selected based on empirical testing through the deployment of the F1Tenth platform in the real world. The chosen configuration, with 6 convolutional and 3 fully connected layers, achieves a balance between localization accuracy, fast training, and real-time inference performance on the embedded Jetson platform.

    Due to their repetitive and uniform geometries, tracks pose an ambiguity challenge to CNNs in localization tasks. 
    This leads to similar LiDAR scans at different points of the track, typically long straights or repetitive curves.
    To mitigate this issue, we integrate a positional encoding of the previous pose estimate into the network.
    The encoding initially projects the raw pose ($x, y, \theta$) of the previous timestep into a higher-dimensional space via sine and cosine functions.
    Directly using the previous pose as an input can lead to high sensitivity regarding the input position and overfitting.
    Therefore, we discretize the previous pose into a grid representation, dividing each pose value by the grid cell size and rounding the result to the closest value. 
    Through this process, the network learns a robust spatial connection between two poses and avoids the CNN ambiguity problem.
 
    The loss function for the localization module is defined as the Mean Squared Error between the predicted and ground truth (GT) global pose:
    \begin{equation}
    \mathcal{L_{\text{loc}}} = \frac{1}{3} \left( \left( \hat{x} - x \right)^2 + \left( \hat{y} - y \right)^2 + \left( \hat{\theta} - \theta \right)^2 \right),
    \label{eq:localization_loss}
    \end{equation}
    where \( \mathcal{L_{\text{loc}}} \) is the localization loss, \( \hat{x}, \hat{y}, \hat{\theta} \) are the predicted, and \( x, y, \theta \) the ground truth coordinates and orientation.


    We generate LiDAR-scan to global pose combinations by extracting pose samples from the recorded map and simulating the corresponding scans, which are used to train the localization module in a supervised manner solely on the localization loss.
    Static environment mapping is performed with the \textit{slam\_toolbox} \cite{macenskiSLAMToolboxSLAM2021} deployed directly on the F1Tenth vehicle, creating a Portable Gray Map occupancy grid.
    Additional maps are of the same data format and created by extracting the track boundaries from real-world racetracks, followed by scaling them to match the dimensions of the F1Tenth platform \cite{Betz2022_RacingSurvey}.
    Section \ref{sec:test_scenarios} presents the scenarios in more detail.
    
    An algorithm uses the track's map image to create a tensor of $x, y$ positions from the white pixels representing the driveable area. 
    The orientation is chosen randomly between $0$ and $2  \pi$.
    A raytracing algorithm creates the corresponding LiDAR scan, which is stored as an additional tensor.
    The algorithm adds random noise to the initial $x$ and $y$ positions, so the poses are not evenly spaced on the track according to the pixel discretization.
    Gaussian noise with a mean of $0$, and a standard deviation of $0.25$ is added to the LiDAR scan.
    The value for the standard deviation is derived by deploying models with different values on the real vehicle and evaluating the performance in the real world. 
    As shown in Zang et al. \cite{zangLocal_INNImplicitMap2022}, generating the previous pose during training by adding noise around the ground truth pose is sufficient. 
    During inference, the real previous pose is used. 
    The process is repeated for every white pixel, creating an initial dataset. 
    We further enhance it, by extending the dataset 10-fold, with different noise for the poses.
    We then train a separate localization network for each scenario on the generated dataset using the localization loss.
    In addition, after training the localization network in a supervised manner for 50 epochs solely on the localization loss, we rerun the dataset generator to receive a new scan-pose dataset with new noise, improving generalisability.


    \subsection{Lateral control module}
    \label{sec:control_module}

    Based on the pose estimate from the localization module, the Stanley Controller guides the vehicle along predefined waypoints.

    The Stanley Controller first identifies the closest waypoint to the vehicle's current pose. 
    It then calculates the cross-track error (CTE) $e_{\text{cross}}$ and heading error (HE) $e_{\text{head}}$  relative to that waypoint. 
    The resulting control law for the Stanley Controller is defined as:
    \begin{equation}
    \delta =k_h\;e_{\text{head}} + \arctan\left(\frac{k_e\; e_{\text{cross}}}{v}\right),
    \end{equation}
    where \( \delta \) is the steering angle, \( e_{\text{head}} \) the heading error, \( e_{\text{cross}} \) the cross-track error, and \( v \) the vehicle's speed. The tuning parameter \( k_e \) adjusts the controller’s sensitivity based on the cross-track error, and \( k_h \) is the tuning parameter for the heading error. 
    The parameters \( k_e \) and \( k_h \) are hand-tuned to \( 1.8 \) and \( 1.3 \) for Sections \ref{sec:control_loss} to \ref{sec:real_world}. 
    Section \ref{sec:combined_training} demonstrates the ability to train the parameters.

    \subsection{End-to-End training - Differentiability}
    \label{sec:end-to-end-training}

    At this point, the localization module and Stanley controller operate independently. The localization module is only trained on the localization loss \( \mathcal{L_{\text{loc}}} \), and the Stanley controller follows the predefined waypoints according to its control law based on the pose estimate. 
    Now, the end-to-end differentiability comes into play, where we extend the role of the localization module beyond simple state estimation, enabling it to reduce the control loss for the specific controller. 
    This optimization is only made possible through the end-to-end differentiable MIND-Stack architecture.

    The new loss function we train the localization module against combines three individual loss terms. The first is the squared cross-track error, a well-established error metric quantifying how well a vehicle follows its reference trajectory by representing both longitudinal and lateral deviations.

    To calculate the cross-track error, we need the vehicle's current pose. 
    We address this by solving another problem the localization model presents.
    Training the network to output an optimal pose estimate for the controller, not necessarily the ground truth position, hinders user interpretability since the user does not know if the output reflects the real pose or is optimized for the control error. 
    To overcome this and as input for the error calculation, we always run an unaltered localization module parallel to the one we train. 
    We, therefore, have an accurate pose estimate and further increase the user's system understanding. 
    They can directly derive where the vehicle is and can compare it to the output of the localization network trained toward the control loss, providing them with a sense of understanding of how the training changes the model's estimate. 

    To connect the error metric to the control module, we utilize a kinematic bicycle model \cite{althoffCommonRoadComposableBenchmarks2017}, propagating it two timesteps of \SI{0.01}{s} into the future and using the resulting pose estimate as input for the cross-track error calculation. 
    This ensures that the error depends on the Stanley's steering output, enabling gradient flow from the error function through the controller to the localization module.

    The final loss function combines the cross-track error, a term penalizing large changes in the vehicle's orientation, and the original localization loss:
    \begin{equation}
    \mathcal{L_{\text{total}}} =  \alpha \; e_{\text{cross}}^2 + \beta \mid \theta_{t} - 2 \; \theta_{t-1} + \; \theta_{t-2} \mid + \gamma \; \mathcal{L_{\text{loc}}},
    \label{eq:total_loss}
    \end{equation}
    
    where \(\alpha\), \(\beta\), and \(\gamma\) are factors adjusting the influence of each term loss to the total loss.

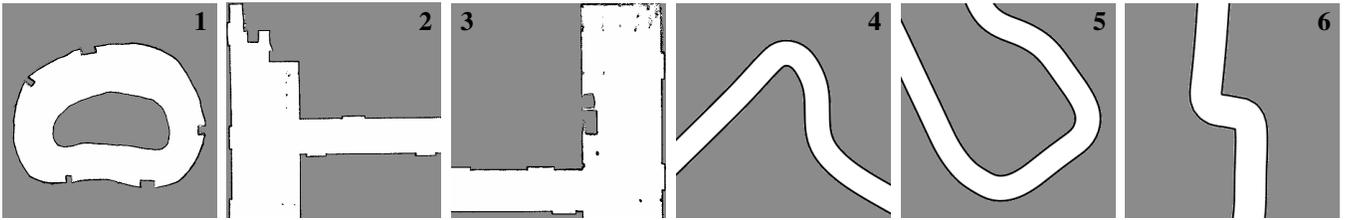
\begin{figure*}[t!]
        \centering
        \begin{minipage}[t]{0.16\textwidth}
            \centering
            \begin{tikzpicture}
                \node[anchor=south west,inner sep=0] (image) at (0,0) {\includegraphics[width=\textwidth]{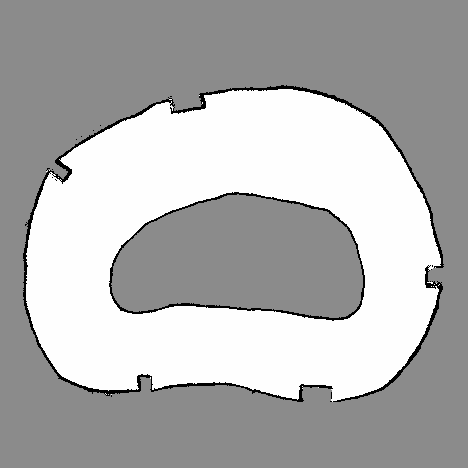}};
                \node[anchor=north east, text=black, font=\bfseries] at (image.north east) {1};
            \end{tikzpicture}
        \end{minipage}
        \hfill
        \begin{minipage}[t]{0.16\textwidth}
            \centering
            \begin{tikzpicture}
                \node[anchor=south west,inner sep=0] (image) at (0,0) {\includegraphics[width=\textwidth]{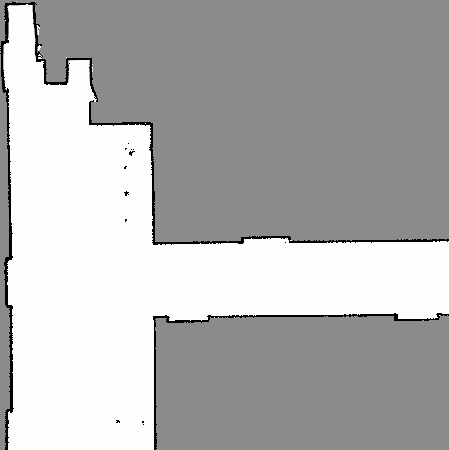}};
                \node[anchor=north east, text=black, font=\bfseries] at (image.north east) {2};
            \end{tikzpicture}
        \end{minipage}
        \hfill
        \begin{minipage}[t]{0.16\textwidth}
            \centering
            \begin{tikzpicture}
                \node[anchor=south west,inner sep=0] (image) at (0,0) {\includegraphics[width=\textwidth]{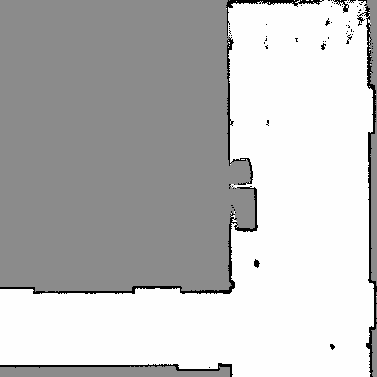}};
                \node[anchor=north west, text=black, font=\bfseries] at (image.north west) {3};
            \end{tikzpicture}
        \end{minipage}
        \hfill
        \begin{minipage}[t]{0.16\textwidth}
            \centering
            \begin{tikzpicture}
                \node[anchor=south west,inner sep=0] (image) at (0,0) {\includegraphics[width=\textwidth]{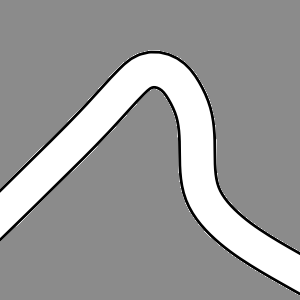}};
                \node[anchor=north east, text=black, font=\bfseries] at (image.north east) {4};
            \end{tikzpicture}
        \end{minipage}
        \hfill
        \begin{minipage}[t]{0.16\textwidth}
            \centering
            \begin{tikzpicture}
                \node[anchor=south west,inner sep=0] (image) at (0,0) {\includegraphics[width=\textwidth]{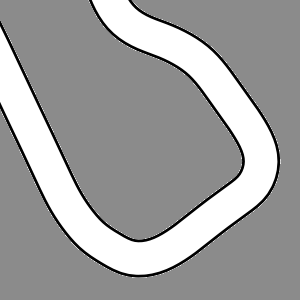}};
                \node[anchor=north east, text=black, font=\bfseries] at (image.north east) {5};
            \end{tikzpicture}
        \end{minipage}
        \hfill
        \begin{minipage}[t]{0.16\textwidth}
            \centering
            \begin{tikzpicture}
                \node[anchor=south west,inner sep=0] (image) at (0,0) {\includegraphics[width=\textwidth]{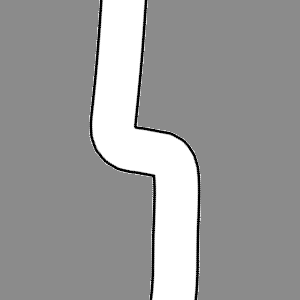}};
                \node[anchor=north east, text=black, font=\bfseries] at (image.north east) {6};
            \end{tikzpicture}
        \end{minipage}
        \caption{Six scenarios used to train and evaluate MIND-Stack, presenting different challenges: From left to right Scenario 1 - Scenario 6. }
        \label{fig:six_images}
    \end{figure*}

    \subsection{Modularity and interpretability}
    \label{sec:mod-and-interpret}

    MIND-Stack has a modular structure that retains the task subdivision with distinct, replaceable modules. 
    In regards to the presented stack, each module can be simply exchanged for any other differentiable localization or control algorithm, utilizing the same interfaces. 
    Additional modules, enabling new applications can be added in between, as long as the downstream to upstream gradient flow is not interrupted.
    MIND-Stack's structure enables these new algorithms to directly optimize their performance towards the final driving task. 
    Each intermediate result in MIND-Stack, such as the pose from the localization module, or the cross-track and heading error provided by the control algorithm, corresponds to human interpretable physical variables. 
    In addition, hand-crafted differentiable algorithms such as the Stanley Controller remain fully interpretable as their entire inner behavior consists of mathematical equations. 
    This enables MIND-Stack to remain interpretable as traditional rule-based approaches and improve over full neural network architectures. 

\section{EXPERIMENTAL SETUP}

We conduct simulation and real-world experiments to evaluate MIND-Stack's performance. 
    The following section details the utilized hardware and software, as well as testing scenarios.

    \subsection{Real-world autonomous driving platform}
    \label{sec:platform_sim}
    
    The real-world experiments are performed on a widespread autonomous driving platform called \textit{F1Tenth} \cite{okellyF1TenthOpensourceEvaluation2020}. 
    This autonomous vehicle is based on the Traxxas Slash 4x4 chassis, features a 2D Hokuyo UST-10LX LiDAR Sensor, and an NVIDIA Jetson Orin Nano CPU/ GPU, operating in the 15 W power mode as an embedded system.
    The simulation experiments are conducted in a simulation environment called \textit{F1Tenth Gym}, running at a fixed sample time of \SI{0.01}{s} (\SI{100}{Hz}) on a workstation with an NVIDIA GTX 4090 GPU and an AMD Ryzen 9 7950X CPU. 
    All simulation-based experiments reported in Section~\ref{sec:results} use this fixed timestep to compute the evaluation metrics.
    Both the vehicle and workstation offer CUDA cores and, therefore, compatibility with PyTorch and GPU acceleration.

    \subsection{Test scenarios}
    \label{sec:test_scenarios}
    
    We utilize different scenarios to train and evaluate MIND-Stack's performance (Figure \ref{fig:six_images}). 
    The environments consist of 2 segments and 1 track from the real world (Scenarios 1-3), as well as 3 segments from real racetracks  (Scenarios 4 and 5: Hockenheimring, Scenario 6: Circuito de Barcelona-Cataluña), scaled down to the F1Tenth platform's scale.
    The scenarios represent challenging track features, including hairpin corners, sharp bends, and fast corners. 
    We conduct each experiment on a per-scenario basis and include evaluation before and after training. 
    Results are averaged over 30 evaluation laps per phase while training on the control loss is performed for 30 laps per scenario with specific hyperparameters derived via manual grid search (Table \ref{tab:hyperparameter_overview}). Further details are provided in Section~\ref{sec:control_loss}.

    \begin{table}[h]
    \centering
    \caption{Hyperparameter settings for each scenario (\(lr\): learning rate). All experiments are trained for 30 laps.}
    \begin{tabular}{l c c c c c}
    \toprule
    \textbf{Experiment (Section)} & $lr$ & $\alpha$ & $\beta$ & $\gamma$ \\
    \midrule
    Scenario 1 (\ref{sec:control_loss}, \ref{sec:comparison}, \ref{sec:combined_training}) 
      & \(9 \times 10^{-8}\) & 5.5 & 1.0 & -- \\ 
    Scenario 2 (\ref{sec:control_loss}, \ref{sec:real_world}) 
      & \(4 \times 10^{-8}\) & 5.5 & -- & -- \\
    Scenario 3 (\ref{sec:control_loss}, \ref{sec:real_world}) 
      & \(4 \times 10^{-8}\) & 5.5 & 1.0 & -- \\
    Scenario 4 (\ref{sec:control_loss}) 
      & \(3 \times 10^{-9}\) & 1.0 & -- & 0.005 \\
    Scenario 5 (\ref{sec:control_loss}) 
      & \(3 \times 10^{-9}\) & 5.5 & 1.0 & -- \\
    Scenario 6 (\ref{sec:control_loss}) 
      & \(8 \times 10^{-9}\) & 5.5 & -- & -- \\
    Stanley (\ref{sec:combined_training}) 
      & \(1 \times 10^{-3}\) & 5.5 & -- & -- \\
    \bottomrule
    \end{tabular}
    \label{tab:hyperparameter_overview}
\end{table}

\section{RESULTS}
\label{sec:results}
We present the experimental results of MIND-Stack in both simulation and real-world scenarios. 
    To our knowledge, these results showcase the first implementation of a fully differentiable software stack, covering the entire pipeline from sensor input to drive output, incorporating both neural networks and hand-crafted algorithms into one framework. 
    Unlike prior research, which either misses a direct connection to the sensors and actuators or fails to apply their solution to the real world, our work demonstrates both capabilities. 
    This section shows MIND-Stack’s ability to optimize upstream modules on downstream control losses through end-to-end differentiability, achieving improved results compared to state-of-the-art algorithms, running on a real-world autonomous platform, and training both the localization and control module together for increased performance.
    All results in this section are presented as mean values per simulation timestep to ensure comparability across different scenarios.

    \subsection{MIND-Stack optimizes the control loss}
    \label{sec:control_loss}

    \begin{table*}[!t]
    \centering
    \caption{Comparison of the training loss and validation loss before and after optimizing MIND-Stack towards the control loss for each scenario. The training loss represents the mean value of Equation \ref{eq:total_loss} per simulation timestep over 30 laps, while the validation loss is the mean absolute cross-track error in regard to the vehicle's ground truth pose per timestep.}
    \begin{tabular}{c c c | c c}
        \toprule
        & \multicolumn{2}{c|}{Before Optimization} & \multicolumn{2}{c}{\textbf{After Optimization}} \\
        \cmidrule(lr){2-3} \cmidrule(lr){4-5}
        Scenario & Training Loss & Validation Loss & Training Loss & Validation Loss \\
        \midrule
        1 & 39.26 ± 1.58 & 25.83 cm ± 0.91 & 32.37 ± 1.21 & 23.25 cm ± 0.42 \\
        2 & 49.09 ± 1.97 & 30.66 cm ± 0.58 & 16.36 ± 1.27 & 12.15 cm ± 1.95 \\
        3 & 12.96 ± 1.21 & 9.04 cm ± 0.64 & 5.21 ± 0.22 & 5.83 cm ± 0.27 \\
        4 & 7.66 ± 0.27 & 8.87 cm ± 0.20 & 4.49 ± 0.27 & 6.19 cm ± 0.29 \\
        5 & 1.62 ± 0.20 & 12.05 cm ± 0.82 & 0.12 ± 0.03 & 2.72 cm ± 0.47 \\
        6 & 5.95 ± 0.70 & 8.99 cm ± 0.61 & 1.68 ± 0.09 & 3.58 cm ± 0.27 \\
        \bottomrule
    \end{tabular}
    \label{tab:cross_track_scenarios}
\end{table*}

    \begin{figure}[b]
        \centering
        \input{Plots/Sim_Crosstrack_Loss/sim_crosstrack_loss}
    \end{figure}

    \begin{figure}[b!]
    \begin{tikzpicture}
    \node[anchor=south west,inner sep=0] (image) at (0,0) {
        \input{Images/raceline_image/real_racetrack_racelines}
    };
    \begin{scope}[x={(image.south east)},y={(image.north west)}]
        \node[fill=white,draw=none,anchor=north west, text width=4.8cm, inner sep=2pt] at (0.02,1.04) {
            \renewcommand{\arraystretch}{0.9}
            \begin{tabular}{@{}rl@{}}
                \tikz \node[draw, fill=Black, circle, minimum size=4pt, inner sep=0pt] at (0,0) {}; & \hspace{0.0em}Reference Waypoints

                \\
                \textcolor{Bluelight}{\rule{2em}{2pt}} & Traj. before optimization \\
                \textcolor{Orange}{\rule{2em}{2pt}} & Traj. after optimization
            \end{tabular}
        };
    \end{scope}
    \end{tikzpicture}
    \caption{Visualization of the new driven trajectory (\textcolor{Orange}{orange}) after optimization compared to the original trajectory (\textcolor{Bluelight}{blue}) before optimization in Scenario 1.}
    \label{fig:raceline_visualization}
    \end{figure}
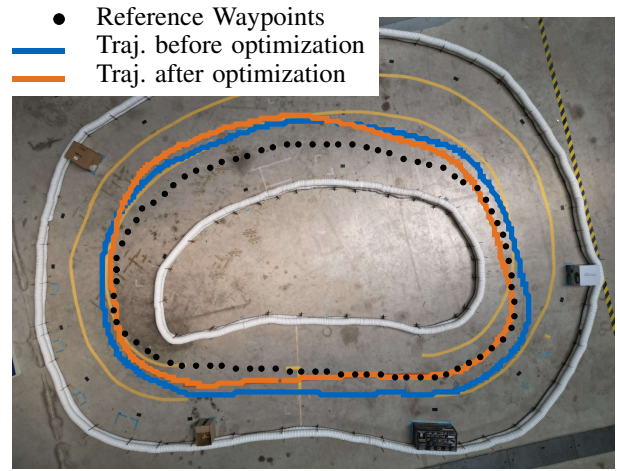

    In order to clearly present our findings, we first define consistent metrics and recurring terms. 
    Except for the state-of-the-art comparison in Section \ref{sec:comparison}, all evaluation metrics include both a training and validation loss. 
    The training loss is defined in Equation \ref{eq:total_loss} and refers to the mean loss per simulation timestep over one lap.  
    We use the mean absolute cross-track error per timestep with respect to the vehicle's ground truth pose as our validation loss.
    This measure reflects how well an algorithm is able to follow its trajectory.
    The per-timestep formulation ensures consistent and comparable evaluation across scenarios with different lap durations and simulation step counts.

    Our experiments show the improvements of using end-to-end differentiability during training. 
    We define two phases: Before Optimization and After Optimization. 
    Before Optimization refers to the phase where the localization network is only trained on the localization loss for the current scenario (Equation \ref{eq:localization_loss}), without end-to-end training. 
    We evaluate this model for 30 laps in its respective scenario without further training to gather statistically significant baseline results. The results differ between laps, as factors such as varying LiDAR noise influence the vehicle's behavior and lead to variations in its trajectory.
    We then train the dedicated localization network for each scenario for 30 laps on the control loss, extending its role beyond state estimation.
    The resulting model for the scenario is used in the After Optimization phase to evaluate the stack using the localization model optimized toward the control loss. 
    It is evaluated for 30 laps in the same scenario without further training.
    The goal is to show that training and validation losses decrease from Before to After Optimization, demonstrating the improvement stemming from the end-to-end differentiability and training.
    We calculate the mean and standard deviation per timestep and scenario and report the results in Table \ref{tab:cross_track_scenarios}.
 
    Figure \ref{fig:crosstrack_error} demonstrates how the training process continuously reduces the training loss over each training lap in Scenario 2.
    Figure \ref{fig:raceline_visualization} visualizes the trajectory the vehicle drives before optimization (blue), not using end-to-end differentiability but solely a localization module trained on the localization loss and after optimization (orange). 
    It is evident that the vehicle is able to improve its path-tracking capabilities through its end-to-end differentiability and associated training.    
    

    We show MIND-Stack's ability to work and generalize in all environments, reducing the training and validation loss for all six scenarios from before to after optimization (Table \ref{tab:cross_track_scenarios}).
    MIND-Stack reaches a minimum training loss reduction of \SI{17.55}{\%} for Scenario 1 and maximum of \SI{92.59}{\%} for Scenario 5. 
    The average improvement across all scenarios is \SI{58.29}{\%}.
    Similarly, end-to-end training reduces the validation loss from at least \SI{9.99}{\%} up to \SI{77.43}{\%}. The average reduction is \SI{45.61}{\%}.


    \subsection{MIND-Stack outperforms state-of-the-art algorithms} 
    \label{sec:comparison}
    
    By comparing MIND-Stack’s performance against state-of-the-art algorithms for autonomous navigation in \mbox{Scenario 1}, we assess its effectiveness and advantages in relation to established approaches. 
    
    We compare the validation loss of MIND-Stack before (solely trained on the localization loss) and after optimization (trained for 30 laps on the control loss) to state-of-the-art algorithms \cite{evansUnifyingF1TenthAutonomous2024a} such as Follow-the-Gap (FTG) \cite{sezerNovelObstacleAvoidance2012}, Pure Pursuit (PP) \cite{Coulter-1992-13338}, Stanley Controller (SC) \cite{hoffmannAutonomousAutomobileTrajectory2007}, and Model Predictive Control (MPC).
    PP, SC, and MPC use a Particle Filter (PF) as localization input \cite{walshCDDTFastApproximate2018b}. 
    We utilize the Particle Filter for PP, SC, and MPC as it is the most common localization combined with these algorithms.

\begin{table}[b]
    \centering
    \caption{Comparison of MIND-Stack against state-of-the-art localization and control algorithms in Scenario 1. \textbf{MIND-Stack offers the best performance} by at least \SI{11.93}{\%} to the other algorithms via its end-to-end performance.}
    \begin{tabular}{lcc}
        \toprule
        Algorithm & Validation Loss & Lap Time \\
        \midrule
        Follow-the-gap & 100.03 cm $\pm$ 2.4 & 9.22 s $\pm$ 0.07 \\
        PF + Pure Pursuit & 26.4 cm $\pm$ 0.37 & 7.41 s $\pm$ 0.02 \\
        PF + Stanley Controller & 31.60 cm $\pm$ 0.92 & 7.63 s $\pm$ 0.03 \\
        PF + MPC & 58.47 cm $\pm$ 0.72 & 4.97 s $\pm$ 0.01 \\
        MIND (before optim.) & 25.83 cm $\pm$ 0.91 & 5.33 s $\pm$ 0.01 \\
        \textbf{MIND (after optim.)} & \textbf{23.25 cm} $\mathbf{\pm}$ \textbf{0.42} & 5.03 s $\pm$ 0.02 \\
        \bottomrule
    \end{tabular}
    \label{tab:mind_comparison}
\end{table}

    Table \ref{tab:mind_comparison} showcases the validation loss accumulated over one lap and lap time of each algorithm. 
    Even before optimization, MIND-Stack has a lower validation loss compared to the other algorithms and shows its superior performance, further reducing the error by \SI{9.99}{\%} using the model optimized towards the control loss (after optimization). 
    All at competitive speeds and lap times, compared to the state-of-the-art algorithms.

    \subsection{MIND-Stack transfers to the real world}
    \label{sec:real_world}

    MIND-Stack is evaluated in two challenging real-world scenarios (Scenario 2 and 3) at speeds of up to \SI{4.27}{m/s}, identical to the simulation environment.
    The tests are crucial to demonstrate that MIND-Stack can be transferred from simulation to real-world deployment. The real-world experiments operate at a sample time of \SI{0.025}{s} (\SI{40}{Hz}), set by the update rate of the Hokuyo LiDAR sensor, in comparison to the simulation environment (Section \ref{sec:platform_sim}).
    We first evaluate the algorithm before optimization for 10 laps on the real vehicle without further training. Then, we run a model previously trained in the simulation to minimize the control loss for 10 laps without any further training on the real vehicle.
    Through the efficient implementation of the MIND-Stack framework, designed specifically for the F1Tenth's embedded low-power NVIDIA Jetson compute platform, the run-time is very fast at \SI{0.019}{s}, with a faster possible frequency at over \SI{50}{Hz} than the LiDAR input data.
    The network directly uses the real LiDAR scan as input and outputs the steering command to the vehicle.

\newcolumntype{R}[1]{>{\raggedleft\arraybackslash}p{#1}}

\begin{table}[b]
    \centering
    \caption{Comparison of the training loss and validation loss in the real-world experiments before and after optimizing MIND-Stack towards the control loss.}
    \begin{tabular}{lR{1.3cm}R{1.3cm}|R{1.3cm}R{1.3cm}}
        \toprule
        & \multicolumn{2}{c|}{Before Optimization} & \multicolumn{2}{c}{\textbf{After Optimization}} \\
        \cmidrule(lr){2-3} \cmidrule(lr){4-5}
        Scenario & Training Loss & Validation Loss & Training Loss & Validation Loss \\
        \midrule
        2 & 43.12 $\pm$ 3.52 & 30.1 cm $\pm$ 3.6 & 30.93 $\pm$ 2.09 & 10.0 cm $\pm$ 1.39 \\
        3 & 44.33 $\pm$ 6.23 & 20.23 cm $\pm$ 5.29 & 24.88 $\pm$ 2.51 & 9.96 cm $\pm$ 0.44 \\
        \bottomrule
    \end{tabular}
    \label{tab:real_world_results}
\end{table}

    Table \ref{tab:real_world_results} shows the mean loss and ground truth values per timestep before and after training towards the control loss. 
    It is clear that in both scenarios, MIND-Stack is able to transfer the improved performance from the simulation environment to the real-world and real autonomous platform, achieving a statistically significant reduction in training loss and validation loss.
    MIND-Stack achieves a reduction of up to \SI{43.87}{\%} in the training loss and \SI{50.77}{\%} in the validation loss.

    \subsection{MIND-Stack enables combined localization and Stanley Controller training}
    \label{sec:combined_training}

\begin{table}[t]
    \centering
    \caption{Comparison of the training loss and validation loss across different training setups in Scenario 1, \textbf{showing the benefits of combined localization and control module training}.}
    \begin{tabular}{p{3.0cm}p{2.0cm}p{2.1cm}} 
        \toprule
        Training Setup & Training Loss & Validation Loss \\
        \midrule
        Localization Only & 39.26 $\pm$ 1.58 & 25.83 cm $\pm$ 0.91 \\
        After Control Loss Optimization & 32.37 $\pm$ 1.21 & 23.25 cm $\pm$ 0.42 \\
        Stanley Controller Only & 27.91 $\pm$ 0.88 & 22.67 cm $\pm$ 2.65 \\
        \textbf{Combined Training (Control Loss + Stanley)} & \textbf{21.33 $\pm$ 2.11} & \textbf{19.95 cm $\pm$ 1.6} \\
        \bottomrule
    \end{tabular}
    \label{tab:combined_training_results}
\end{table}

    Only through MIND-Stack's unique architecture, enabling gradient propagation throughout the entire software stack, can we unlock the full potential of end-to-end differentiable algorithms: The simultaneous training of both the localization network and Stanley Controller's parameters \( k_e \) and \( k_h \), all toward the control loss.
    The advantages are illustrated through a comparison of different training setups. 
    The first setting (Localization Only) serves as the baseline, training solely towards minimizing the localization loss and equals the before optimization state from the previous sections.
    The second approach, "After Control Loss Optimization," represents the state after optimization, utilizing the end-to-end differentiability and extending the task of the localization module to control loss minimization.
    The third setting, "Stanley Controller Only," exclusively trains the Stanley Controller parameters isolated from any other module.

    While both the second and third approaches show improvements over the baseline (\SI{17.55}{\%} for control loss optimization and \SI{28.91}{\%} for Stanley training in regards to the training loss), they each do not reach the performance that results from simultaneous optimization presented in the fourth setting, with both the localization network and Stanley Controller training towards the control loss together \mbox{(Table \ref{tab:combined_training_results})}.
    This setting demonstrates the full potential of MIND-Stack: a framework where upstream or downstream modules are trained simultaneously against the same downstream loss function. 
    This reduces the mean training loss per timestep by \SI{23.58}{\%} and the mean validation loss per timestep by \SI{12.00}{\%} compared to the second-best approach.
    A neural network (Localization) and a hand-crafted rule-based algorithm (Stanley) train together and work together to further improve the vehicle's capabilities, which is only possible through end-to-end differentiability.

\section{SUMMARY \& FUTURE WORK}

This paper introduced MIND-Stack, a modular, interpretable, and end-to-end differentiable software stack designed for autonomous navigation utilizing real-world autonomous systems and embedded hardware.
Our approach integrates a lightweight localization module with a traditional Stanley Controller. 
The end-to-end differentiability enables the upstream localization module to be trained on the downstream control loss.
Unlike previous research that lacks direct sensor and actuator interfaces or real-world validation, MIND-Stack comprehensively demonstrates its capabilities in both simulation and real-world environments. 
We demonstrate its robustness in different challenging racetrack and real-world scenarios, improving the mean cross-track error per timestep by up to \SI{77.43}{\%} using its full end-to-end differentiability.
MIND-Stack outperforms state-of-the-art algorithms by at least \SI{11.93}{\%} regarding their path-following capabilities.
Real-world tests showcase simulation to real-world transferability and its efficient performance using limited computational resources on an embedded autonomous platform.
All these benefits are achieved while the software stack remains interpretable through intermediate state representations and modular to integrate new algorithms in the future, offering a significant advancement over traditional rule-based or full neural network-based end-to-end algorithms.

We showcased the capabilities of MIND-Stack in using its end-to-end differentiability to train both the localization module and controller simultaneously in a path tracking setting. 
The next step is, therefore, to extend MIND-Stack to all modules of the autonomous driving stack – perception, prediction, planning, and control.
Stable algorithm design has proven to significantly enhance the training process, making it interesting to integrate more intelligent but also more complex localization and control algorithms, such as MPC, into the framework.
Utilizing the added capabilities it will be possible to deploy MIND-Stack to more complicated scenarios with dynamic obstacle interactions.
This creates a fully differentiable framework where neural networks and hand-crafted algorithms work together and train together to elevate the capabilities beyond previously known performance limitations.

\section*{ACKNOWLEDGMENT}

The authors would like to thank Benjamin Evans, Mattia Piccinini, Rainer Trauth, Zirui Zang, and Baha Zarrouki for providing their valuable feedback and Mingyan Zhou for providing the MPC benchmark algorithm.

\bibliographystyle{IEEEtran}
\bibliography{literatur}

\end{document}

%% file: Plots/Sim_Heading_Loss_Squared/sim_heading_loss_squared.tex
    \begin{tikzpicture}
        \begin{axis}[
            width=1.2\textwidth, 
            height=1.1\textwidth, 
            xlabel={Training Lap},
            xmin=0, xmax=30,
            ymin=0, ymax=0.5,
            xtick={0,10,20,30},
            ytick={0,0.1,0.2,0.3,0.4,0.5},
            grid=major,
            grid style={lightgray!50},
            legend pos=north east,
            legend style={
            at={(0.5,1.1)}, 
            anchor=south,
            draw=none,
            cells={align=left}
            },
            every axis plot/.append style={thick},
            tick align=outside,
            enlargelimits=false
        ]
        \addplot[
            color=Bluelight,
            mark=*,
            mark options={fill=Bluelight},
            line width=0.75pt,
            mark size=2pt,
            nodes near coords,
            point meta=explicit symbolic
        ] table [x=Step, y=Value2, col sep=comma] {Plots/Sim_Heading_Loss_Squared/SHLS.csv};

        \legend{Mean Squared Cross-Track\\ Error per Timestep in \unit{m^2}}
        \end{axis}
    \end{tikzpicture}

%% file: Plots/Sim_Crosstrack_Loss/sim_crosstrack_loss.tex
    \begin{tikzpicture}
        \begin{axis}[
            width=0.4\textwidth, 
            height=0.25\textwidth, 
            xlabel={Training Lap},
            ylabel={Sq. CTE \(e^2_{\text{cross}}\) in \unit{m^2}},
            xmin=0, xmax=30,
            ymin=0, ymax=0.5,
            xtick={0,10,20,30},
            ytick={0,0.1,0.2, 0.3,0.4,0.5},
            grid=major,
            grid style={lightgray!50},
            legend pos=north east,
            legend style={draw=none, cells={align=left}},
            every axis plot/.append style={thick},
            tick align=outside,
            enlargelimits=false
        ]
        \addplot[
            color=Bluelight,
            mark=*,
            mark options={fill=Bluelight},
            line width=0.75pt,
            mark size=2pt,
            nodes near coords,
            point meta=explicit symbolic
        ] table [x=Step, y=Value2, col sep=comma] {Plots/Sim_Heading_Loss_Squared/SHLS.csv};

        \legend{Mean Sq. Cross-Track \\ Error per Timestep}
        \end{axis}
    \end{tikzpicture}
    \caption{Training the localization module to minimize the control error in Scenario 2, shows a clear \textbf{reduction in the mean control loss per timestep}.}
    \label{fig:crosstrack_error}

%% file: Images/raceline_image/real_racetrack_racelines.tex
\pgfplotsset{compat=newest}
\usetikzlibrary{plotmarks}

\begin{tikzpicture}
    \node[anchor=south west,inner sep=0] (image) at (0,0) {\includegraphics[width=0.45\textwidth]{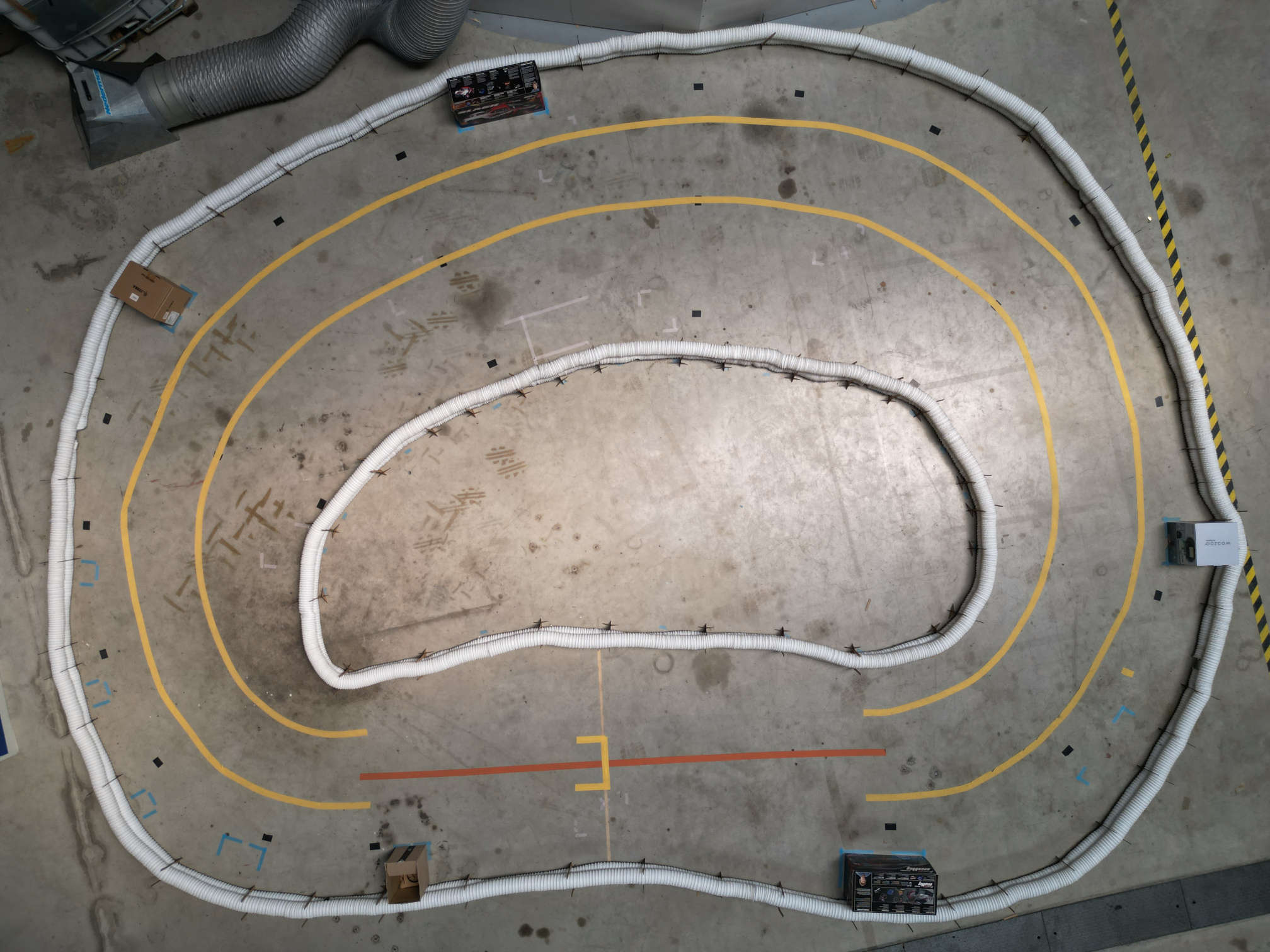}};
    \begin{scope}[x={(image.south east)},y={(image.north west)}]
        

        \begin{axis}[
            at={(image.south west)},
            anchor=south west,
            width=\textwidth,
            height=\textheight,
            axis lines=none,
            clip=false,
            xmin=0, xmax=\pgfkeysvalueof{/pgfplots/width},
            ymin=0, ymax=\pgfkeysvalueof{/pgfplots/height}
        ]
            \addplot [
                mark=none,
                Bluelight,
                line width=2pt,
            ] table [x=x, y=y, col sep=comma] {Images/raceline_image/ConditionalCNNmin305.csv};
            \addplot [
                only marks,
                mark=*,
                mark size = 1pt,
                Black,
            ] table [x=x, y=y, col sep=comma] {Images/raceline_image/waypoints.csv};
            \addplot [
                mark=none,
                Orange,
                line width=2pt,
            ] table [x=x, y=y, col sep=comma] {Images/raceline_image/16_27_20_5.csv};
        \end{axis}
    \end{scope}
\end{tikzpicture}